\def\year{2020}\relax
\documentclass[letterpaper]{article} 
\usepackage{aaai20}  
\usepackage{times}  
\usepackage{helvet} 
\usepackage{courier}  
\usepackage[hyphens]{url}  
\usepackage{graphicx} 
\usepackage{graphicx}  
\usepackage{xcolor}
\usepackage{bm}

\usepackage{multirow}
\usepackage{algorithm}
\usepackage{algpseudocode}
\usepackage{amsmath}

\usepackage{xspace}

\usepackage{hyperref} 

\makeatletter
\DeclareRobustCommand\onedot{\futurelet\@let@token\@onedot}
\def\@onedot{\ifx\@let@token.\else.\null\fi\xspace}

\def\eg{\emph{e.g}\onedot} 
\def\ie{\emph{i.e}\onedot} 
 
\def\etc{\emph{etc}\onedot} 
 
\def\etal{\emph{et al}\onedot}
\makeatother

\usepackage{amsfonts}
\newcommand{\algname}{\textsc{L-GCN }}
\newcommand{\algnamens}{\textsc{L-GCN}}

\definecolor{hdcolor}{RGB}{39, 174, 96}
\def\hd{\textcolor{hdcolor}}

\def\qi{\textcolor{black}}

\title{Location-aware Graph Convolutional Networks for Video Question Answering }

\author{
    Deng Huang\textsuperscript{\rm 1}\thanks{ Equal contribution.}, 
    Peihao Chen\textsuperscript{\rm 1}\footnotemark[1], 
    Runhao Zeng\textsuperscript{\rm 1},
    Qing Du\textsuperscript{\rm 1},
    Mingkui Tan\textsuperscript{\rm 1,\rm 2}\thanks{ Corresponding author.},
    Chuang Gan\textsuperscript{\rm 3} \\
    \textsuperscript{\rm 1}South China University of Technology,
    \textsuperscript{\rm 2}Peng Cheng Laboratory, Shenzhen,
    \textsuperscript{\rm 3}MIT-IBM Watson AI Lab \\
    sehuangdeng@mail.scut.edu.cn,
    \{duqing, mingkuitan\}@scut.edu.cn,
    \{phchencs, runhaozeng.cs, ganchuang1990\}@gmail.com
}

\begin{document}
	
	\maketitle
	
	\begin{abstract}

        We addressed the challenging task of video question answering, which requires machines to answer questions about videos in a natural language form. Previous state-of-the-art methods attempt to apply spatio-temporal attention mechanism on video frame features without explicitly modeling the location and relations among object interaction occurred in videos. However, the relations between object interaction and their location information are very critical for both action recognition and question reasoning.  In this work, we propose to represent the contents in the video as a \textit{location-aware} graph by incorporating the location information of an object into the graph construction. Here, each node is associated with an object represented by its appearance and location features. Based on the constructed graph, we propose to use graph convolution to infer both the category and temporal locations of an action. 
		As the graph is built on objects, our method is able to focus on the foreground action contents for better video question answering.  Lastly, we leverage an attention mechanism to combine the output of graph convolution and encoded question features for final answer reasoning.
		Extensive experiments demonstrate the effectiveness of the proposed methods. Specifically, our method significantly outperforms state-of-the-art methods on TGIF-QA, Youtube2Text-QA and MSVD-QA datasets. Code and pre-trained models are publicly available at: 
        \url{https://github.com/SunDoge/L-GCN}

	\end{abstract}

	\section{Introduction}

\begin{figure}[t]
		\centering
		\includegraphics[width = 1\linewidth]{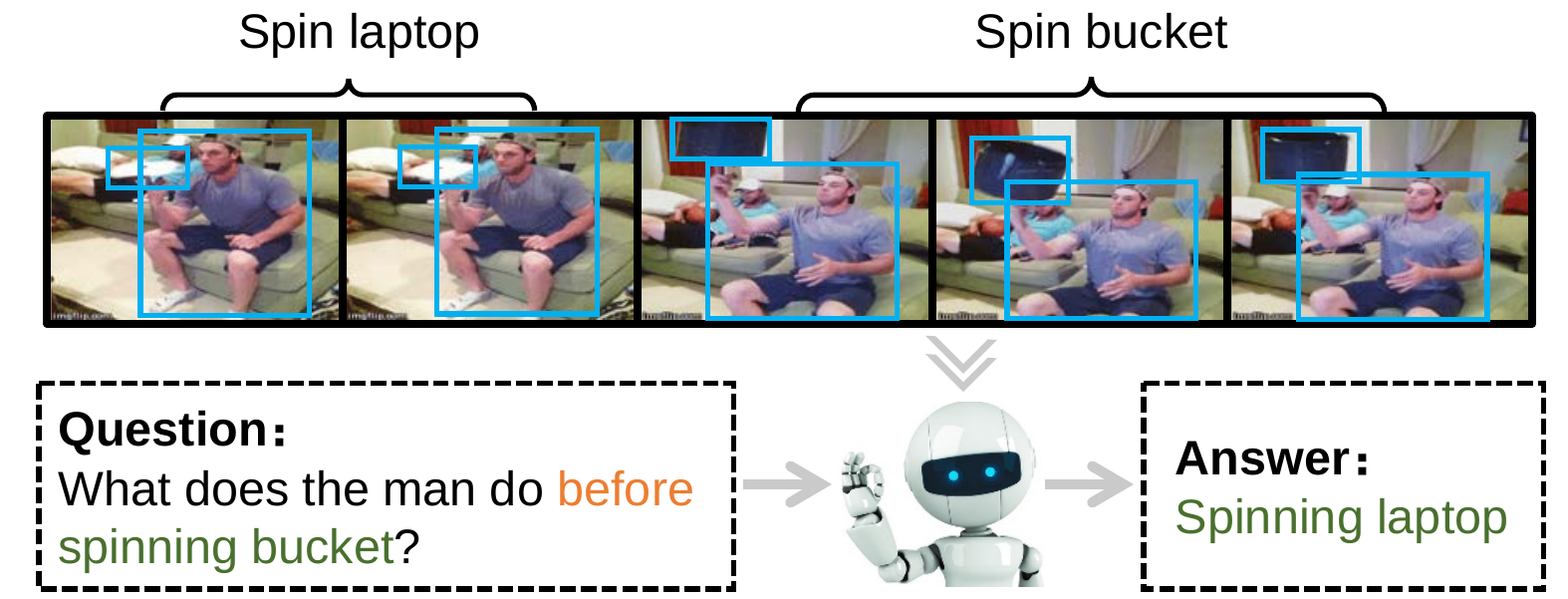}
		
		\caption{One question-answer (QA) pair in video QA task. To answer the question, the model is required to recognize actions (labeled in green) from the interaction between objects (labeled in blue boxes) and be aware of their temporal order (\eg, \textit{before}).}
		\label{fig:sample}
	\end{figure}

	Recently, deep learning has witnessed a great process \cite{JMLR-tan,cao2018adversarial,cao2019multi,Gan_2019_ICCV,guo2019auto}.
	Video question answering (video QA) has become an emerging task in computer vision and has drawn increasing interests over the past few years due to its vast potential applications in artificial question answering system and robot dialogue, video retrieval, etc. In this task, a robot is required to answer a question after watching a video.
	Unlike the well-studied Image Question Answering (image QA) task which focuses on understanding static images~\cite{bottom-up,pythia,DMN}, video QA is more practical since the input visual information often change dynamically, as shown in Figure \ref{fig:sample}.

	Compared with image QA, video QA is much more challenging due to several reasons. (1) Visual content is more complex in a video since it may contain thousands of frames, as shown in Figure \ref{fig:sample}. More importantly, some frames may be dominated with strong background content which however is irrelevant to questions. (2) Videos often contain multiple actions, but only a part of them are of interest to questions.
	(3) Questions in video QA task often contain queries related to temporal cues, which implies we should consider both temporal location of objects and complex  interaction between them for answer reasoning. For example in Figure \ref{fig:sample}, to answer the question ``What does the man do \textbf{\textit{before}} spinning bucket?", the robot should not only recognize the actions ``spin laptop'' and ``spin bucket'' by understanding the interaction between the man and objects (\ie, laptop and bucket) in different frames, but also find out the temporal order of actions (\eg,  \textit{before/after}) for answer reasoning along time axis.

	Taking video frames as inputs, most existing methods \cite{HME,PSAC} employ some spatio-temporal attention mechanism on frame features to ask the network ``where and when to look''.
	However, these methods are often not robust due to complex background content in videos.
	Lei \textit{et al.} \cite{tvqa} tackle this problem by detecting the objects in each frame and then processing the sequence of object features via an LSTM. However, the order of the input object sequence, which may affect the performance, is difficult to arrange.
	More importantly, processing the objects in a recurrent manner will inevitably neglect the direct interaction between nonadjacent objects. This is critical for video QA (see experiments in Section \ref{sec:ablation}).
	
	In this paper, we introduce a simple yet powerful network named Location-aware Graph Convolutional Networks (\algnamens) to model the interaction between objects related to questions. We propose to represent the content in a video as a graph and identify actions through graph convolution.
	Specifically, the objects of interest are first detected by an off-the-shelf object detector. Then, we construct a fully-connected graph where each node is an object and the edges between nodes represent their relationship. 
	We further incorporate both spatial and temporal object location information into each node, letting the graph be aware of the object locations. When performing graph convolution on the object graph, the objects directly interact with each other by passing message through edges. Last, the output of GCNs and question features are fed into a visual-question interaction module to predict  a answer. 
	Extensive experiments demonstrate the effectiveness of the proposed location-aware graph. We achieve state-of-the-art results on TGIF-QA, Youtube2Text-QA and MSVD-QA datasets.
	
	The main contributions of the proposed method are as follows: (1) we propose to explore actions for video QA task through learning interaction between detected objects such that irrelevant background content can be explicitly excluded; (2) we propose to model the relationships between objects through GCNs such that all objects are able to interact with each other directly; (3) we propose to incorporate object location information into graph such that the network is aware of the location of a specific action; (4) our method achieves state-of-the-art performance on TGIF-QA, Youtube2Text-QA and MSVD-QA datasets.

	\section{Related Work}
	
	
	\textbf{Visual Question Answering (VQA)}
	is a task to answer the given question based on the input visual information. 

	Based on the visual sources, we can classify the VQA tasks into two categories: image QA \cite{balanced_vqa_v2,gan2017vqs} and video QA \cite{tvqa,yi2019clevrer}.
	Image QA focuses on spatial information. Most image QA models adopt attention mechanism to capture spatial area that related to question words. 
	Yang \etal \cite{SAN} proposed a multi-layer Stacked Attention Network (SAN) which uses  questions as query to extract the image region related to the answer. 
	Anderson \etal \cite{bottom-up} combined bottom-up and top-down attention which connect questions to specific objects detected by Faster-RCNN. After that, associating feature vector with visual regions becomes a popular framework in the VQA research (\ie Pythia \cite{pythia}).
	Xiong \etal \cite{DMN} introduced the dynamic memory network (DMN) architecture to image QA, which strengthens the reasoning ability of network. 


    In video QA task, understanding untrimmed videos~\cite{zeng2019breaking,wu2019multi} is important. 
    To this end, Jang \etal \cite{tgifqa} utilized both motion (i.e. C3D) and appearance (\ie, ResNet \cite{resnet}) features to better represent the video.
	Li \etal \cite{PSAC} replaced RNN with self-attention together with location encoding to model long-range dependencies.
	However, all the existing methods neglect the interaction between objects, which is vital for video QA task.
	
	\textbf{Graph-based reasoning}
	has been popular in recent years \cite{zeng2019graph,guo2019nat} and shown to be powerful for relation reasoning. 
	To dynamically learn graph structures, CGM \cite{mk-tan} applied a cutting plane algorithm to iteratively activate a group of cliques.
	Recently, Graph Convolution Networks (GCNs) \cite{gcn} have been used for semi-supervised classification.
	In text-based tasks, such as machine translation and sequence tagging, GCNs breaks the sequence restriction between each word and learns the graph weight by attention mechanism, which makes it work better in modeling longer sequence than LSTM. Some methods \cite{conditionedgraph,murel,regat} took into consideration the object position for image QA tasks. In video recognition, Wang \textit{et al.} \cite{JointGCN} proposed to use GCNs to capture relations between objects in videos, where objects are detected by an object detector pre-trained on extra training data. Despite their success, there is no efficient graph model for video QA task.
	
	\textbf{Attention mechanism}
	has been leveraged in various tasks. Several works \cite{gan2015devnet,long2018attention} used attention model to improve the performance on video recognition. Vaswani \textit{et al.} \cite{attention} utilized self-attention mechanism for language translation and \cite{co-attention} proposed Co-Attention which can be stacked to form a hierarchy for multi-step interactions between visual and language features. Jang \etal \cite{tgifqa} proposed a simple baseline which uses both spatial and temporal attention to reason the video and answer the question. In our 
	proposed
	method, we use attention mechanism to fuse video and question modalities.
	
	\begin{figure*}[ht]
		\centering
		\includegraphics[width=\textwidth]{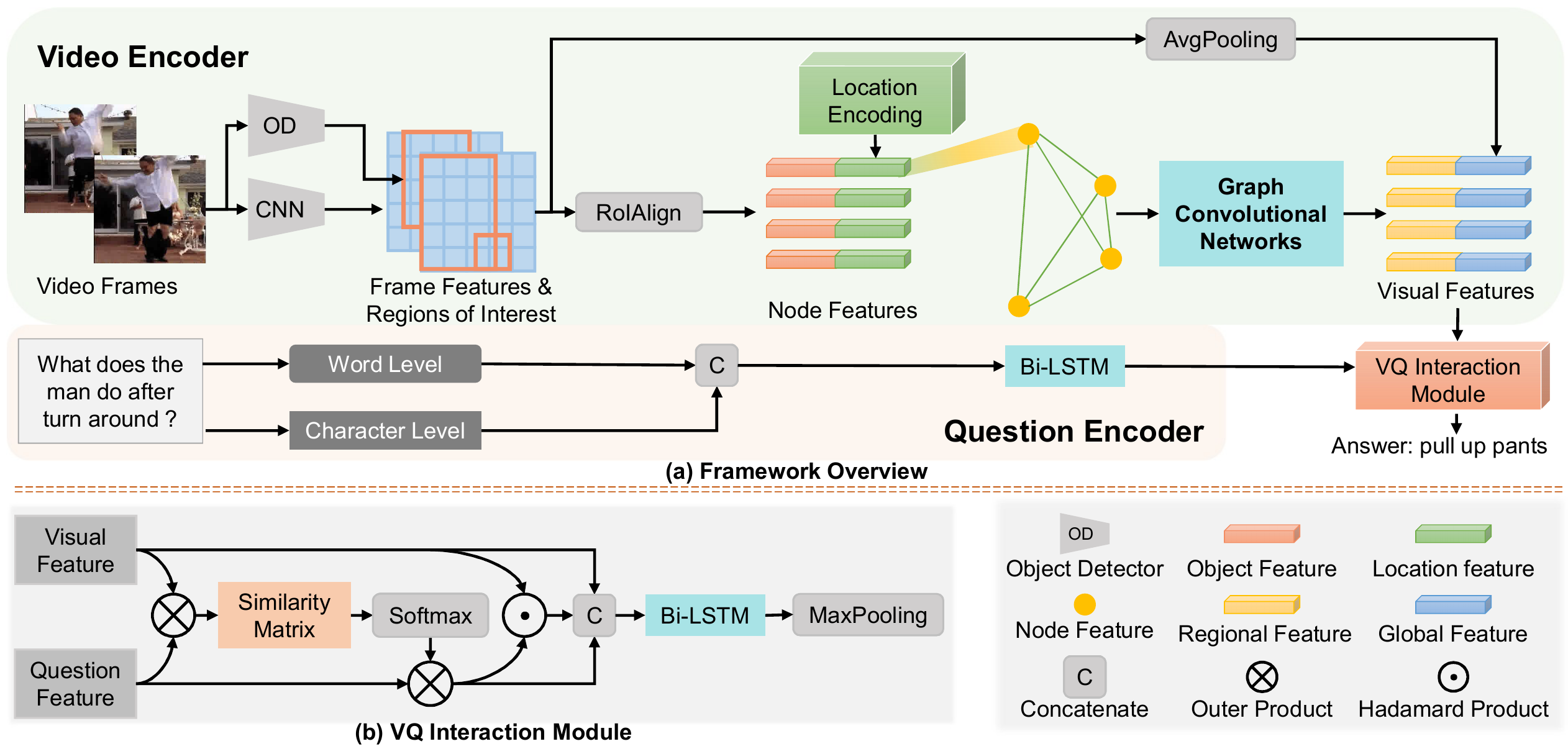}
		\caption{Illustration of the proposed method. \algname consists of two streams, namely the \textbf{question encoder} stream and the \textbf{video encoder} stream, which process queries and video contents, respectively.  The outputs of two streams are combined with a \textbf{visual-question (VQ) interaction module}. The \textbf{location-aware graph} built on objects considers both interactions of objects and their temporal location information.}
		\label{fig:model-overview}
	\end{figure*}

	\section{Proposed Method}
	\subsection{Notation and Problem Definition}
	
	Given a video containing $N$ frames with $K$ detected objects on each frame, let $\mathcal{R}{=}\{\mathbf{o}_{n,k}, \mathbf{b}_{n,k}\}_{n=1, k=1}^{n=N, k=K}$ be the detected object set, where $\mathbf{o}$ denotes the object feature obtained by RoIAlign \cite{mask-rcnn} and $\mathbf{b}$ is the spatial location
	of each object. We use $T{=}N{\times} K$ to denote the total number of objects in one video.
	We denote a graph as $\mathcal{G} = (\mathcal{V}, \mathcal{E})$ with $M$ nodes $\mathbf{v}_i \in \mathcal{V}$ and edges $e_{ij} \in \mathcal{E}$. The adjacency matrix of graph is represented as $\mathbf{A} \in \mathbb{R}^{M \times M}$. The question with $\kappa$ words is denoted as $\mathrm{Q}$.

	In this paper, we focus on video QA task, which requires the model to answer questions related to a video.
	This task is challenging as video contents are complex with strong irrelevant backgrounds. Besides, most QA pairs in video QA task are related to more than one action with temporal cues.
	To answer the question correctly, the model is required not only to recognize the actions correctly from complex contents but also to be aware of their temporal order.
	
	\subsection{General Scheme}

    The general scheme of our method is shown in Figure \ref{fig:model-overview}, which consists of two streams. 
	The first stream is regarding a {question encoder}, which processes queries with a Bi-LSTM. The second stream is related to a {video encoder}, which focuses on understanding video contents by exploiting a \textbf{location-aware graph} built on objects.   The outputs of two streams are then combined  by a {visual-question (VQ) interaction module}, which employs an attention mechanism to explore which question words are more relevant to the visual representation. Last, the answer is predicted by applying an FC layer on top of the VQ interaction module. 
	
\begin{algorithm}[th]
	\caption{Overall training process.}
	\begin{flushleft}
		\textbf{Input:}
		Video frame features; 
		object set $\mathcal{R}$;
		question $\mathrm{Q}$
	\end{flushleft}
	\begin{algorithmic}[1]
		\State Construct the location-aware graph $\mathcal{G}$ as in Section~\ref{sec:graph}
		\While {not converges}

		\State Extract question features $\mathbf{F}^Q$ via Eq. (\ref{eq:fq})
		\State Encode object location via Eq. (\ref{eq:spatial_location}), (\ref{eq:temporal_location-1}) and (\ref{eq:temporal_location-2}) 
		\State Compute the node features via Eq. (\ref{eq:node_ft})
		\State Update adjacent matrix via Eq. (\ref{eq:adj})
		\State Perform reasoning on graph via Eq. (\ref{eq:graph_con})
		\State Obtain visual features $\mathbf{F}^V$ via Eq. (\ref{eq:fv})
		\State Obtain $\mathbf{F}^C$ from $\mathbf{F}^V$ and $\mathbf{F}^Q$ via Eq. (\ref{eq:interact})
		\State Predict answers from $\mathbf{F}^C$ with answer predictor

		\EndWhile
	\end{algorithmic}
	\begin{flushleft}
		\textbf{Output:}
		Trained model for video QA
	\end{flushleft}
	\label{alg:training}
\end{algorithm}

    In this paper, the \textbf{location-aware graph} plays a critical role. Specifically, we use an object graph $\mathcal{G} {=} (\mathcal{V}, \mathcal{E})$ to model the relationships between objects in a video. Note the temporal ordering of actions in the video is important for  answer reasoning w.r.t. a question in a video QA task. We thus propose to integrate the spatial and temporal location information into the object features  of each node in a graph (See details  in Section~\ref{sec:graph}). In this way, we can exploit both spatial and temporal order information of actions for temporally related answer reasoning.
	
	For convenience, we present the overall training process in Algorithm \ref{alg:training}. In the following, we first describe the \textbf{question encoder}. Then we depict the construction of the location-aware graph and the graph convolution for message passing, followed by description of \textbf{visual encoder}.
	After that, we detail the \textbf{visual-question interaction module}. Last, we present the answer reasoning and loss functions.

	\subsection{Question Encoder Stream} \label{sec:question_encoder}
	
	Given a question sentence, the question encoder is to model the question for video QA.
	To handle the out-of-vocabulary words as well as the misspelling words, we apply both character embedding $\mathbf{Q}^c \in \mathbb{R}^{\kappa \times c \times d_c}$ and word embedding $\mathbf{Q}^w \in \mathbb{R}^{\kappa \times d_w}$ to represent a question $\mathrm{Q}$ with $\kappa$ words, where $d_c$ and $d_w$ denote the dimensions of character embedding and word embedding, respectively. 
	
	In the optimization, the word embedding function is initialized with a pre-trained 300-dimension GloVe \cite{Glove}, and the character embedding function is randomly initialized.
	Given the character and word embeddings, the question embedding can be represented by a two-layer highway network $h(\cdot, \cdot)$ \cite{highway}, which is proven to be effective to solve the training difficulties, that is: 

	\begin{equation}\label{eq:fq}
	\mathbf{Q} = h(\mathbf{Q}^w , g(\mathbf{Q}^c)),
	\end{equation} 
	where the character embedding is further processed by a $g(\cdot)$ which consists of a 2D convolutional layer. 
	
	To better encode the question, we feed the question embedding $\mathbf{Q}$ into a bi-directional LSTM (Bi-LSTM).
	Then we obtain the question feature $\mathbf{F}^Q$ by stacking the hidden states of the Bi-LSTM from both directions at each time step. 

	\subsection{Location-aware Graph Construction}\label{sec:graph}
	
Given a video with $K$ detected objects for each frame, we seek to represent the video into a graph. Noting that actions can be inferred from the interaction between objects, we thus construct a fully-connected graph on the detected objects.  
We may use object features to represent each node. However, this node type ignores the location information of objects, which is vital for temporally related answer reasoning.
To address this, we will describe how to encode the location information with so-called location features. 
With location features, we are able to construct a location-aware graph, namely, we concatenate both object appearance and location features as node features.

	\paragraph{Location Encoding.}
	Given a detected object in the $n^{th}$ frame with spatial location $\mathbf{b}$ and aligned feature $\mathbf{o}$, we encode its spatial location feature $\mathbf{d}^s$ with a Multilayer Perceptron ($\mathrm{MLP}(\cdot)$) which consists of two FC layers and a ReLU activation function \cite{relu}, that is:
	\begin{equation} \label{eq:spatial_location}
	\mathbf{d}^s = \mathrm{MLP}(\mathbf{b}),
	\end{equation} 
	where $\mathbf{b}$ is represented by the top-left coordinate and the width and the height of detected objects.
	
	Moreover, we also encoder temporal location feature $\mathbf{d}^t$ of objects using sine and cosine functions of different frequencies \cite{attention} as follows:
	\begin{equation} \label{eq:temporal_location-1}
	{d}_{2j}^t = \mathrm{sin}(n/10000^{2j/d_p}),
	\end{equation} 
	\begin{equation} \label{eq:temporal_location-2}
	{d}^t_{2j+1} = \mathrm{cos}(n/10000^{2j/d_p}),
	\end{equation} 
	where $d_{i}^t$ is the $i$-th entry of the temporal location feature $\mathbf{d}^t$, and $d_p$ is its dimension.
	Then, the feature of each graph node can be defined as:
	\begin{equation} \label{eq:node_ft}
	\mathbf{v} = [\mathbf{o}; \mathbf{d}^s; \mathbf{d}^t],
	\end{equation} 
	where $[\cdot; \cdot; \cdot]$ concatenates three vectors into a longer vector. In this way, each node in the graph contains not only the object appearance features but also the location information.

\subsection{Reasoning with Graph Convolution}\label{sec:g_conv}

Given the constructed location-aware graph, we perform graph convolution to obtain the regional features.
In our implementation, we build $P$-layer graph convolutions. Specifically, for the $p$-th layer (1 $\leq$ $p$  $\leq$ $P$), the graph convolution can be formally represented as:
\begin{equation} \label{eq:graph_con}
\mathbf{X}^{(p)} = \mathbf{A}^{(p)} \mathbf{X}^{(p-1)} \mathbf{W}^{(p)},
\end{equation} 
where $\mathbf{X}^{(p)}$ is the hidden features of the $p$-th layer;  $\mathbf{X}^{(0)} $ is the input node features $\mathbf{v}$ in Eq. (\ref{eq:node_ft});  $\mathbf{A}^{(p)}$ is the adjacency matrix calculated from the node features in the $p$-th layer; and $\mathbf{W}^{(p)}$ is the trainable weight matrix.
Let $\mathbf{X}^{(P)}$ be the output of the last layer of the $P$-layer GCNs. 
Then, we define the regional features $\mathbf{F}^{R}$ as:
%
\begin{equation}
\mathbf{F}^{R} = \mathbf{X}^{(P)} + \mathbf{X}^{(0)}.
\end{equation} 
This can be considered as a skip connection of input  $\mathbf{X}^{(0)}$ and output $\mathbf{X}^{(P)}$, and it  helps to improve the training performance, similar to ResNet~\cite{resnet}.
In our method, the adjacency matrix is a learnable matrix, which is able to simultaneously infer a graph by learning the weight of all edges. 
We calculate the adjacency matrix by:
\begin{equation}\label{eq:adj}
\mathbf{A}^{(p)} = \mathrm{softmax}\left(\mathbf{X}^{(p-1)} \mathbf{W}_1 \cdot (\mathbf{X}^{(p-1)} \mathbf{W}_2)^{\sf{T}}\right),
\end{equation}
where $\mathbf{W}_1$ and $\mathbf{W}_2$ are projection matrices. The softmax operation is performed in the row axis.

	\subsection{Visual Encoder Stream}
	The visual encoder is to model video contents via object interaction for video QA.
	Given a $N$-frame video, we extract frame features using a fixed feature extractor (\eg,  ResNet-152). At the same time, $K$ bounding boxes are detected for each frame by an off-the-shelf object detector. The object features $\mathbf{o}$ are obtained using RoIAlign \cite{mask-rcnn} on top of the image features, followed by an FC layer and ELU activation function \cite{elu} to reduce dimension.

	Given the detected object set $\mathcal{R}$, we construct a location-aware graph $\mathcal{G}(\mathcal{V}, \mathcal{E})$ on the objects. Then, we perform graph convolution to enable the message passing between objects through edges, which can be formally represented as:
	\begin{equation}
	\mathrm{GCN}(\mathcal{G}(\mathcal{V}, \mathcal{E}), \{[\mathbf{o}_{t}; f(\mathbf{b}_{t})]\}_{t=1}^{T}),
	\end{equation}
	where ${[\cdot; \cdot]}$ indicates the concatenation of vectors and $f(\cdot)$ denotes for any mapping function, \eg, multi-layer perceptron (MLP). 
	The output of GCNs is termed as regional features $\mathbf{F}^R$. 
	Besides, in order to introduce the context information, we apply global average pooling on the frame features to generate global features $\mathbf{F}^G$. 
	
	The global features are further processed by a 1D convolutional layer and an ELU activation function to merge the information from neighbor frames. After that, we replicate the global features $K$ times
	and employ Multilayer Perceptron (MLP) (with one hidden layer and an $\mathrm{ELU}$ activation function) to merge the concatenation of $\mathbf{F}^R$ and $\mathbf{F}^G$, which yields visual features $\mathbf{F}^V$ and that is:
	\begin{equation}\label{eq:fv}
	\mathbf{F}^V = \mathrm{MLP}([\mathbf{F}^R , \mathbf{F}^G]).
	\end{equation} 


	\subsection{Visual-question Interaction Module} \label{sec:vq-interaction}
	After obtaining visual and question representations, we propose a visual-question (VQ) interaction module to combine them for predicting answer. The framework of VQ interaction module is shown in Figure \ref{fig:model-overview}(b). We first map $\mathbf{F}^V$ and $\mathbf{F}^Q$ into the same subspace with dimension $d_s$ through two independent FC layers, leading to  $\mathbf{F}^V \in \mathbb{R}^{T \times d_s}$ and $\mathbf{F}^Q \in \mathbb{R}^{L\times d_s}$.
	Then, we explore which question words are more relevant to each visual representation for video QA. In this paper, we leverage attention mechanism to learn a cross modality representation inspired by \cite{BiDAF}.
	
	Specifically, we first calculate similarity matrix $\mathbf{S}$ between $\mathbf{F}^V$ and $\mathbf{F}^Q$ via dot product together with a softmax function applying along each row, that is: 
	\begin{equation}
	\mathbf{S} = \mathrm{softmax}\left(\mathbf{F}^V   (\mathbf{F}^Q)^{\sf{T}} \right).
	\end{equation} 
	Then, we calculate the weighted question features $\widetilde{\mathbf{F}}^Q$ corresponding to each visual feature via dot product between $\mathbf{S}$ and $\mathbf{F}^Q$. The cross modality representation $\mathbf{F}^C \in \mathbb{R}^{P \times 3d_s}$ is calculated by:
	\begin{equation}\label{eq:interact}
	\mathbf{F}^C = [\mathbf{F}^V, \widetilde{\mathbf{F}}^Q, \mathbf{F}^V \! \odot \! \widetilde{\mathbf{F}}^Q],
	\end{equation} 
	where $\odot$ means the element-wise product operation. To yield the final representation for answer prediction, we leverage a Bi-LSTM followed by a max pooling layer across the dimension $T$.
	
	\subsection{Answer Reasoning and Loss Function}
	The questions for video QA can be summarized as three types: multiple-choice, open-ended and counting.
	In this subsection, we will describe how to predict answers for each question type given cross modality features $\mathbf{F}^C$.
	
	\textbf{Multiple-choice question}: for this kind of questions, there exist $U$ choices and the model is required to choose the correct one. We first embed the content of each choice in the same way as question encoding described in Section \ref{sec:question_encoder}, leading to $U$ independent answer features $\mathbf{F}^A$. Then, each answer feature is interacted with visual features in the way described in Section \ref{sec:vq-interaction}, where we replace the question feature by answer question, yielding the weighted answer features $\widetilde{\mathbf{F}}^A$. 
	Then, the cross modality representation $\mathbf{F}^C$ in Eq. (\ref{eq:interact}) is constructed as $[\mathbf{F}^V, \widetilde{\mathbf{F}}^Q, \widetilde{\mathbf{F}}^A, \mathbf{F}^V \! \odot \! \widetilde{\mathbf{F}}^Q, \mathbf{F}^V \! \odot \! \widetilde{\mathbf{F}}^A]$. 
	We leverage an identical FC layer on $U$ cross modality representations to predict scores $\mathcal{A} = \{a_1, \ldots, a_U\}$. The scores are processed by a softmax function. We use cross entropy loss as the loss function:
	\begin{equation}
        \begin{array}{ll}
            L_M = -\sum_{i=1}^U y_i \log\left(\frac{e^{a_i}}{\sum_{j=1}^U e^{a_j}}\right),
        \end{array}
	\end{equation}
	where $y_i = 1$ if answer $a_i$ is the right choice, otherwise $y_i = 0$.
	We take the choice with the highest score as the prediction.
	
	\textbf{Open-ended question}: for these questions, the model is required to choose a correct word as answer from the pre-defined answer set of  $C$ candidate words in total. We predict the scores $\mathcal{A} = \{a_1, \ldots, a_{C}\}$ of each candidate word using an FC layer together with a softmax layer. Also, we use the cross entropy loss as the loss function: 
	\begin{equation}
        \begin{array}{ll}
             L_O = -\sum_{i=1}^C y_i \log\left(\frac{e^{a_i}}{\sum_{j=1}^C e^{a_j}}\right),
        \end{array}
	\end{equation}
	where $y_i = 1$ if answer $a_i$ is the right answer, otherwise $y_i = 0$.
We take the word with the highest score as our prediction.
	
	\textbf{Counting question}: for these questions, the model is required to predict a number ranging from 0 to 10. We leverage an FC layer upon $\mathbf{F}^C$ to predict the number. We use mean square error loss to train the model: 
	\begin{equation}
        L_C = \left\| \mathbf{x}-\mathbf{y} \right\|_2^2,
	\end{equation}
	where $\mathbf{x}$ is the predicted number, $\mathbf{y}$ is the ground truth.
	During the testing, the prediction is rounded to the nearest integer and clipped within 0 to 10.

	\begin{table*}[ht]
		\centering
			\caption{Statistics of three video QA datasets.  \#MC denotes the number of options for multiple-choice questions.}
		\scalebox{0.95}{
		\begin{tabular}{c|ccccccc}
			\hline
			Dataset         & Vocab\hd{.} size & \#Video & \#Question & Answer size & \#MC & Feature type   & \#Sampled frame \\ \hline
			TGIF-QA         & 8,000      & 71,741   & 165,165     & 1,746       & 5                  & ResNet-152     & 35               \\
			Youtube2Text-QA & 6,500      & 1,970    & 99,429      & 1,000       & 4                  & ResNet-101+C3D & 40               \\ 
			MSVD-QA         & 4,000      & 1,970    & 50,505      & 1,000       & NA                 & VGG+C3D        & 20               \\ \hline
		\end{tabular}
		}
		\label{table:datasets}
	\end{table*}

	\section{Experiments}
	In this section, we first introduce three benchmark datasets and implementation details. Then, we compare the performance of our model with the state-of-the-art methods. Last, we perform ablation studies to understand the effect of each component.

	\subsection{Datasets}
	We evaluate our method on three video QA datasets. The statistics of the datasets are listed in Table \ref{table:datasets}. More details are given below. 
	
	\textbf{TGIF-QA} \cite{tgifqa} consists of 165K QA pairs from 72K animated GIFs.
	The QA-pairs are splited into four tasks: 1) Action: a multiple-choice question recognizing action repeated certain times; 2) Transition (Trans.): a multiple-choice question asking about the state transition; 3) FrameQA: an open-ended question that can be inferred from one frame in videos; 4) Count: an open-ended question counting the number of repetition of an action. The multiple-choice questions in this dataset have five options and the open-ended questions are with a pre-defined answer set of size 1,746.
	
	\textbf{Youtube2Text-QA}~\cite{youtube2text-qa-rANL} includes the videos from MSVD video set \cite{msvd} and the question-answer pairs collected from Youtube2Text \cite{youtube2text} video description corpus. It consists of open-ended and multiple-choice questions, which are divided into three types (\ie, \textit{what}, \textit{who} and \textit{others}).
	
	\textbf{MSVD-QA} \cite{msvd-qa} is based on MSVD video set. It consists of five types of questions, including \textit{what}, \textit{who}, \textit{how}, \textit{when} and  \textit{where}. All questions are open-ended with a pre-defined answer set of size 1,000.

	\subsection{Implementation Details}
	
	\paragraph{Evaluation metrics.}
	(1) For the ``Count'' task in TGIF-QA dataset, we adopt the Mean Square Error (MSE) between the predicted answer and the ground truth answer as the evaluation metric.
	(2) For all other tasks in our experiments, we use accuracy to evaluate the performance.

	\paragraph{Training details.}
	We convert all the words in the question and answer to lower cases, and then transform each word to a 300-dimension vector with a pre-trained GloVe model \cite{Glove}.
	For fair comparisons, we adopt the same feature extractors as those are used in the compared methods. More details can be found in Table \ref{table:datasets}.
	We use Mask R-CNN  \cite{mask-rcnn} as object detector and select $K$ detected objects with the highest score for each frame. By default, $K$ is set to 5. The number of GCNs layers is set to 2.
	We employ a Adam optimizer \cite{kingma2014adam} to train the network with an initial learning rate of 1e-4. We set the batch size to 64 and 128 for multiple-choice and open-ended tasks, respectively. 
	

		\subsection{Comparison with State-of-the-art
		Results}

	\paragraph{Results on TGIF-QA.} 
	We compare our \algname with the state-of-the-art methods, including ST-VQA~\cite{tgifqa}, Co-Men~\cite{co-mem}, PSAC~\cite{PSAC} and HME~\cite{HME}.
	From Table \ref{table:tgif-qa},
	our \algname achieves the best performance on four tasks.
	It is worth noting that our method outperforms HME, ST-VQA and Co-Mem by a large margin even if they use additional features (\ie, C3D features~\cite{C3D} and optical flow feature) to model actions. These results demonstrate the effectiveness of leveraging an object graph to capture the object-object interaction and perform reasoning.
	

	\begin{table}[]
		\centering
		\caption{Comparisons with state-of-the-arts on TGIF-QA dataset.  R, C and F denote features extracted by ResNet, C3D and Optical Flow, respectively.}
		\scalebox{0.85}{
			\begin{tabular}{c|cccc}
				\hline
				Model       & Action        & Trans.        & FrameQA       & Count (MSE)         \\ \hline
				ST-VQA(R+C) & 60.8          & 67.1          & 49.3          & 4.28          \\
				Co-Mem(R+F) & 68.2          & 74.3          & 51.5          & 4.10          \\
				PSAC(R)     & 70.4          & 76.9          & 55.7          & 4.27          \\
				HME(R+C)    & 73.9          & 77.8          & 53.8          & 4.02          \\ \hline
				Ours(R)     & \textbf{74.3} & \textbf{81.1} & \textbf{56.3} & \textbf{3.95} \\ \hline
			\end{tabular}
		}
		\label{table:tgif-qa}
	\end{table}

	\paragraph{Results on Youtube2Text-QA.} 
	For further comparison, we test our model on a more challenging dataset Youtube2Text-QA.
	This dataset consists of open-ended and multiple-choice questions, which are divided into three categories (\ie, \textit{what}, \textit{who} and \textit{others}).
		We consider two state-of-the-art baseline methods (HME and r-ANL~\cite{youtube2text-qa-rANL}), and report the results in Table~\ref{table:yout2text-qa}.
	
	
	\qi{From Table~\ref{table:yout2text-qa}, compared with the baselines, our method achieves better performance in overall accuracy in both multi-choice and open-ended questions.
		More specifically, for multiple-choice questions, we achieve the best performance on \textit{what} and \textit{who} tasks. The relatively poor performance on \textit{others} task cannot represent the ability of different models because this kind of questions only occupies 2\% of all QA pairs.
		For open-ended questions, our \algname significantly improves the accuracy from 29.4\% to 53.2\% on \textit{who} task, where most questions are related to the subject of actions. This demonstrates the superiority of leveraging object features, which explicitly localizes the object for video QA task.
	}

	\begin{table}[]
		\centering
		\caption{Comparisons with state-of-the-art methods on Youtube2Text-QA.}
		\scalebox{0.9}{
			\begin{tabular}{c|c|ccc|c}
				\hline
				Task                             & Method & What           & Who            & Other          & All            \\ \hline
				\multirow{3}{*}{Multiple-Choice} & r-ANL  & 63.3          & 36.4          & 84.5          & 52.0          \\
				& HME    & 83.1          & 77.8          & \textbf{86.6}          & 80.8          \\
				& Ours   & \textbf{86.0} & \textbf{81.5} & 80.6 & \textbf{83.9} \\ \hline
				\multirow{3}{*}{Open-Ended}      & r-ANL  & 21.6          & 29.4          & \textbf{80.4} & 26.2          \\
				& HME    & \textbf{29.2} & 28.7          & 77.3          & 30.1          \\
				& Ours   & 24.5          & \textbf{53.2} & 70.4         & \textbf{38.0} \\ \hline
			\end{tabular}
		}
		\label{table:yout2text-qa}
	\end{table}
	
	\paragraph{Results on MSVD-QA.} \
	In Table \ref{table:msvd-qa}, we compare our \algname with ST-VQA, Co-Mem, AMU \cite{msvd-qa} and HME on MSVD-QA dataset. 
	From Table~\ref{table:msvd-qa}, our \algname achieves the most promising performance in overall accuracy, which demonstrates the superiority of the proposed method on the non-trivial scenarios.

	 \begin{table}[]
	 \centering
	 \caption{Comparisons with state-of-the-arts on MSVD-QA.}
	  \scalebox{0.95}{
	 \begin{tabular}{c|cccc|c}
	 \hline
	 Model & ST-VQA & Co-Mem & AMU  & HME  & Ours          \\ \hline
	 Acc   & 31.3   & 31.7   & 32.0 & 33.7 & \textbf{34.3} \\ \hline
	 \end{tabular}
	  }
	 \label{table:msvd-qa}
	 \end{table}

	\subsection{Ablation Study} \label{sec:ablation}

	\paragraph{Impact of each component.}
	We first construct a simple variant of the proposed method as \textbf{baseline}, which uses only the global frame features $\mathbf{F}^G$ to generate visual features $\mathbf{F}^V$ via Eq. (\ref{eq:fv}). Then, the object features, GCNs, and location features will be incorporated into the baseline progressively to generate visual features in higher quality, and we denote them as ``\textbf{OF}'', ``\textbf{GCNs}'' and ``\textbf{Loc}'', respectively. ``\textbf{FC}''and ``\textbf{LSTM}'' represent the models where GCNs are replaced by two Fully-Connected (FC) layers or a  2-layer LSTM, respectively. ``\textbf{Loc\_T}''and ``\textbf{Loc\_S}'' represent the location features which only consist of temporal or spatial location information, respectively. 
	
	We show the results on TGIF-QA dataset in Table \ref{tab:variant}. (1) Compared with the \textbf{baseline}, incorporating object features boosts the performance in all tasks consistently, demonstrating the effectiveness of using detected objects for video QA task. We speculate that the detected objects explicitly help the model exclude irrelevant background. (2) Applying GCNs on object features further boosts the performance, demonstrating the importance of modeling relationships between objects through GCNs. On the other hand, using FC layer or LSTM only brings minor increases or even drops the performance. This is not surprising because the model cannot learn object-object relationship when applying FC layer on each object separately. Besides, objects in different spatial locations cannot be regarded as a sequence and thus LSTM is not suitable for modeling their relationship. (3) Adding location features further increases the performance. Especially, the improvements on the task of transition and count are more significant. One possible reason is that these two tasks are more sensitive to the knowledge of event's order, where the transition task asks about the action transition and the count task asks the number of repetition of an action. We also try to only incorporate temporal or spatial location information into \algnamens. The performance decreases compared to the variant using both location types, demonstrating that these two location information are complementary and both vital for video QA task.

	\paragraph{Impact of \#GCNs layers and detected objects.}
	In this paper, we propose to leverage GCNs on detected objects to learn actions. Here, we conduct ablation studies on the depth of GCNs and the number of the objects in each frame.
	From Table \ref{tab:layer}, GCNs with two layers performs best on three tasks. Considering the efficiency and performance, we leverage 2-layer GCNs by default. Besides, as shown in Table \ref{tab:num_object}, GCNs with 5 detected objects achieves the best performance on three tasks. It is not surprising that the network with 2 detected objects performs worst because the network may neglect some important objects. Additionally, as most of the question answering pairs in TGIF-QA dataset are only relative to a few salient objects, feeding too many objects into network may cripple the performance. By default, we leverage 5 detected objects in experiments.

	\begin{table}[]
		\centering
		\caption{Performance comparisons of different variants on  TGIF-QA. ``\textbf{OF}'' and ``\textbf{Loc}'' denote object  and location features, respectively.}
		\label{tab:variant}
		\scalebox{0.80}{
			\begin{tabular}{l|cccc}
				\hline
				Model                         & Action         & Trans.         & FrameQA        & Count         \\ \hline
				\textbf{baseline}             & 70.58          & 79.59          & 55.37          & 4.33          \\
				\textbf{baseline+OF}          & 72.82          & 80.10          & 55.79          & 4.24          \\
				\textbf{baseline+OF+GCNs}     & 74.10          & 80.39          & 56.10          & 4.15          \\
				\textbf{baseline+OF+GCNs+Loc} & \textbf{74.32} & \textbf{81.13} & \textbf{56.32} & \textbf{3.95} \\
				\hline
				\textbf{baseline+OF+FC}       & 72.96          & 80.18          & 55.94          & 4.22          \\
				\textbf{baseline+OF+LSTM}     & 72.65          & 80.07          & 55.49          & 4.25          \\ \textbf{baseline+OF+GCNs+Loc\_T} & 73.75  & 80.97          & 55.54          & 4.17           \\
				\textbf{baseline+OF+GCNs+Loc\_S} & 73.58  & 80.89          & 56.07          & 4.12           \\
				\hline
			\end{tabular}
		}
		\vspace{-5mm}
	\end{table}

	\begin{table}[]
		\centering
		\caption{Ablation study on \#GCNs layers on TGIF-QA.}
		\label{tab:layer}
		\scalebox{0.90}{
		\begin{tabular}{c|cccc}
			\hline
			\#GCNs layers & Action         & Trans.       & FrameQA        & Count         \\ \hline
			1            & 74.24          & 81.02          & 55.97          & 4.16          \\
			2            & \textbf{74.32} & 81.13          & \textbf{56.32} & \textbf{3.95} \\
			3            & \textbf{74.32} & \textbf{81.58} & 56.23          & 4.16          \\
			4            & 73.97          & 80.86          & 56.01          & 4.10          \\ \hline
		\end{tabular}
		}
	\end{table}

	\begin{table}[]
		\centering
		\caption{Performance comparisons between different numbers of detected objects on TGIF-QA dataset.}
		\label{tab:num_object}
		\scalebox{0.90}{
		\begin{tabular}{c|cccc}
			\hline
			\#objects per frame & Action         & Trans.        & FrameQA        & Count         \\ \hline
			2                 & 74.11          & 80.95          & 55.61          & 4.13          \\
			5                 & \textbf{74.32} & 81.13          & \textbf{56.32} & \textbf{3.95} \\
			10                & 74.01          & \textbf{81.43} & 55.88          & 3.99          \\ \hline
		\end{tabular}
		}
	\end{table}

	\begin{figure}[!t]
		\centering
		\includegraphics[width=0.45\textwidth]{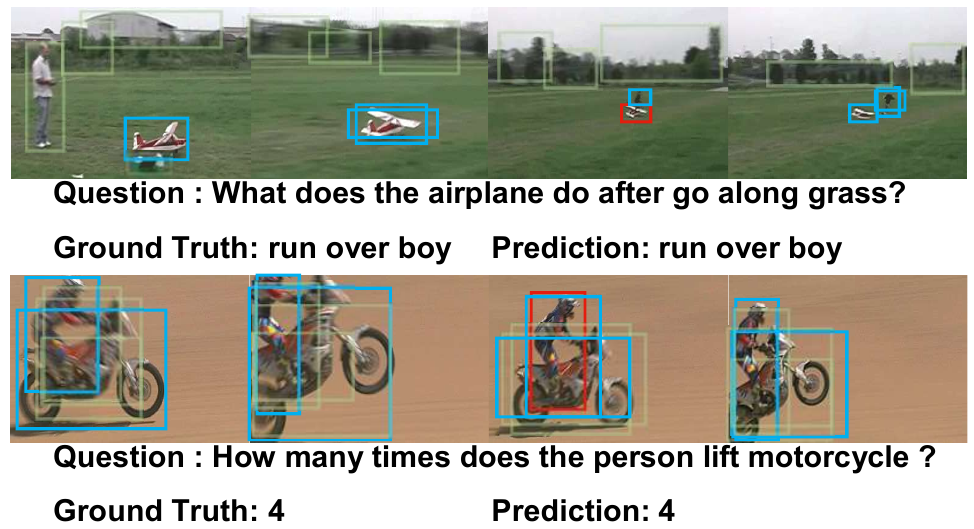}
		\caption{Visualization on TGIF-QA dataset. 
			The boxes in transparent green are $K$ detected objects.
			The boxes in \textcolor{red}{red} are the query object. 
			The boxes in \textcolor{cyan}{blue} are the objects with high values regarding to query object in the adjacent matrix. }
		\label{fig:vis}
	\end{figure}
	
	\subsection{Qualitative Analysis}
	We demonstrate the similarity matrix in the GCNs using two examples in Figure \ref{fig:vis}. 
	We draw two conclusions from these examples: 1) Almost all salient objects which are related to question answering pair have been detected beforehand, such as the airplane and the boy in example 1, the man and the motorcycle in example 2,
    \qi{\etc.}
	These detected objects explicitly help the network avoid the influence from complex irrelevant background content. 2) Our graph not only captures relationships between similar objects in different frames but also focuses on semantic similarity. For the first example, the airplane is correlative to not only itself in different frames but also the little boy. This is helpful to recognize the action of ``airplane running over boy''.

	\section{Conclusion}
	
	In this paper, we have proposed a location-aware graph to model the relationships between detected objects for video QA task. Compared with existing spatio-temporal attention mechanism, \algname is able to explicitly get rid of the influences from irrelevant background content. Moreover, our network is aware of the spatial and temporal location of events, which is important for predicting correct answer. 
	Our method outperforms state-of-the-art techniques on three benchmark datasets.
	   
	\section*{Acknowledgment}
	\begin{footnotesize} 
	This work was partially supported by Guangdong Provincial Scientific and Technological Funds under Grants 2018B010107001, National Natural Science Foundation of China (NSFC) 61602185, key project of NSFC (No. 61836003), Program for Guangdong Introducing Innovative and Entrepreneurial Teams 2017ZT07X183,  Tencent AI Lab Rhino-Bird Focused Research Program (No. JR201902), Natural Science Foundation of Guangdong Province under Grant 2016A030310423, Fundamental Research Funds for the Central Universities D2191240.
	\end{footnotesize}
	
	\bibliographystyle{aaai}
	\small
	\bibliography{AAAI-HuangD.5371}

\end{document}